\documentclass{article}

\usepackage[final]{nips_2016}

\usepackage[utf8]{inputenc} 
\usepackage[T1]{fontenc}    
\usepackage{hyperref}       
\usepackage{url}            
\usepackage{booktabs}       
\usepackage{amsfonts}       
\usepackage{nicefrac}       
\usepackage{microtype}      
\usepackage[pdftex]{graphicx} 

\title{Time Series Prediction: Predicting Stock Price}

\author{
  Aaron Elliot \\
  \texttt{ellioa2@bu.edu} \\
  \And
  Cheng Hua Hsu \\
  \texttt{jack0617@bu.edu} \\
}

\begin{document}

\maketitle

\begin{abstract}
 \hspace{4ex}Time series forecasting is widely used in a multitude of domains. In this paper, we present four models to predict the stock price using the S\&P 500 index as input time series data. The mean (martingale) and ordinary linear models require the strongest assumption in stationarity which we use as baseline models. The generalized linear model (GLM) requires lesser assumptions but is unable to outperform the martingale. In empirical testing, the RNN model performs the best comparing to other two models, because it will update the input through LSTM instantaneously, but also does not beat the martingale. In addition, we introduce an online-to-batch (OTB) algorithm and discrepancy measure to inform readers the state-of-art predicting method, which doesn’t require any stationarity or non-mixing assumptions in time series data. Finally, to apply these forecasting to practice, we introduce basic trading strategies that can create Win-win and Zero-sum situations.
\end{abstract}

\section{Introduction}

\par \hspace{4ex} Time series prediction is a classic problem in many domains, with wide-ranging and high-impact applications. The central problem of forecasting is that of predicting the value \(Y(T+1)\) given past observations \(Y(1), ..., Y(T)\) [3].  Many types of models have been applied to this problem. This paper will present a comparison of four models used for time series forecasting of stock prices.  

\section{Models}

\subsection{Baseline Model}

\par \hspace{4ex} Consider a random variable \(x(t)\) representing the value of a stock at time \(t\geq 0\). That lets us construct the sequence
\(S = \{x(0),x(1),...\}\) of \(x(t)\) in discrete time.
In the analysis of stocks, the sequence of the value is often modeled as a martingale \([1]\). Let's assume that is true, that \b{S} forms a martingale. It follows as a property of martingales that
\begin {equation}\label{eq:1}
\mathbb{E}(x(t)\ |\ x(1),x(2),...,x(t\boldmath{-}1)) = x(t\boldmath{-}1)
\end{equation}
where the best guess for the next value, given all the states up to
now, is where we are now. This can further be generalized to
\begin{equation}\label{eq:2}
\mathbb{E}(x(t)|x(a),x(b),...,x(z))=x(z) \ \ :\ \  a<b<...<z<t.
\end{equation}
the best guess for any value in our sequence, given some sequence of states prior, is the most recent value we know.
\par \hspace{4ex} So, if the sequence of stocks we observe, \(S\), is a martingale, then the best possible algorithm for predicting the next value, is just the current value. Further, as the contrapositive, if we can show a model that  can consistently beat the martingale model, then we have shown there exists dependencies beyond (2).

\par \hspace{4ex} Stocks are also modeled as following a geometric random walk \([2]\). Most famously, this modeling was used in the formulation of the Black-Scholes equation, an equation used throughout the field of asset and risk evaluation \([2]\). This leads to some nice properties over other models, such as the stock's value being always positive, and a stock being worth \$100 or \$1,000,000 not affecting its behavior. 

\par \hspace{4ex} Consider a random variable \(Y(t)\) representing the stock value at continuous time \(t\geq 0\). Let's assume that \(Y(t)\) follows a geometric random walk. As a property, for all \(t_{0}<t_{1}\leq t_{2}<t_{3}\leq...t_{n}<t_{n+1} \), the random variables \(\{ \frac{Y(t_{0})}{Y(t_{1})},\frac{Y(t_{2})}{Y(t_{3})},...,\frac{Y(t_{n})}{Y(t_{n+1})}\}\) are mutually independent. Further, \(\forall \  i,j \ \frac{Y(t_{i})}{Y(t_{j})}\) follows a Log-Normal distribution. This implies that, under the assumption our data is of a geometric random walk, if we transform our data into a sequence of percent change in the stock, then we will have independent data. We will further take advantage of this Log-Normal distribution in our use of a Generalized Linear Model on our data.

\par \hspace{4ex} Linking geometric random walks back to the martingale, if \(\forall \  i,j\) the mean of \( \frac{Y(t_{i})}{Y(t_{j})}\) is 1, then \(\{Y(t)\}^{\infty}_{t=0}\) forms a martingale. For the purpose of our analysis, we will use this as our baseline model, a geometric random walk without drift.
\par \hspace{4ex} Lastly, stocks are often thought to have non-stationary behavior. If an event changes how your variables interact (i.e., Trump tweets, Brexit, etc.) then that is observed as a change in the distribution our random variables are sampled from. All our models assume some form of stationarity, though state of the art models find a way to get around this.

\subsection{Linear Model}

\par \hspace{4ex} Linear models are formulated such that a response variable (y) is a linear combination of a predictor variable, or multiple predictor variables. Linear models assume: (1) the response is normally distributed, (2) the errors are normally distributed and independent, and (3) that the predictors are fixed, with constant variance [5]. Assuming stocks are correctly modeled by a random walk with drift, it follows that
\begin{equation}\label{eq:1}
\mathbb{E}(x(t)\ |\ x(1),x(2),...,x(t\boldmath{-}1)) = x(t\boldmath{-}1) + b
\end{equation}
This behavior is described by the linear model: \(y(t) = \beta x(t-1) + b\).

\subsection{Generalized Linear Model (GLM)} 

\par \hspace{4ex} Generalized linear models extend linear models to allow for the response to  (1) be non-linearly related to the linear combination of predictors, via a link function; (2) follow any distribution in the exponential family (e.g., binomial, poisson, normal, gamma, etc.); and (3) model other types of data (e.g., categorical, ordinal, etc.) [5]. The Maximum Likelihood Estimate of the predictor weights is typically found using a Newton-Raphson method. GLMs assume (1) a linear relation between the response transformed by the link function and the linear combination of predictors, and (2) errors are independent. Homogeneity of variance of the predictors is not assumed.

\par \hspace{4ex} As stock price is often modeled with a log-normal distribution, which has the desirable properties of being positive and non-symmetric, we will be using a GLM with a normal distribution and log link function for this analysis. Assuming stocks are correctly modeled by a geometric random walk, it follows that 
\begin{equation}\label{eq:1}
\mathbb{E}(x(t)\ |\ x(1),x(2),...,x(t\boldmath{-}1)) = \beta x(t\boldmath{-}1)
\end{equation}
This behavior is described by a GLM with log link function and a "memory length" of one sample. The statsmodels python package was used for such modeling [6].

\subsection{Recurrent Neural Network}

\par \hspace{4ex} Recurrent neural networks (RNN) are neural networks which have some way of remembering the previous values inputed into them, and output based on both the input, and the remembered value. This is done by three subprocesses: forget gate, input gate, and output gate. Lets say that the remembered value is an vector \(s\) within the Neural Network. Forget gate is a sub-neural network that, given \(s\), will decide what part of \(s\) to discard from iteration to iteration. The update gate is a sub-neural network that, given an input \(x\) and the current value of \(s\), will output the new value for \(s\). Lastly, the output gate is a sub-neural network which gives the RNN's output as a function \(o(x,s)\).

\par \hspace{4ex} What makes RNN's especially good at predicting time series data is that they allow long term dependencies to be expressed in the output, yet don't need to have complicated architecture to allow for variable size inputs. Further, if there exist long term dependencies in our time series, then a Neural Network given enough nodes can theoretically model them. This is less realistic in practice. Since a neural network is built to follow the gradient descent of the loss function, if a simpler model is a local minima, then it's likely that the model will get stuck in that minima. Further, if an RNN is made too complicated, it may find a very deep minima in its training data, but only because it has overfit. Lastly, a RNN does assume stationarity; since, after the training is finished, the weights of the network are set.
\par \hspace{4ex} Specifically for this analysis, we will be using a Long Short Term Memory recurrent neural network (LSTM-RNN). LSTM-RNN goes a step further to having this memory architecture in its individual nodes. For our architecture we use an input layer connected to a layer of LSTM nodes, followed by a dense layer of 1 node for output. This architecture was settled upon based on initial testing with the hyper-parameters, and based on our data set size of $\approx$4000 entries.

\subsection{State of the Art Predicting Method} 

\par \hspace{4ex} Professors Vitaly Kuznetsov and Mehryar Mohri from Courant Institution give learning guarantees for regret minimization algorithms for forecasting nonstationary non-mixing time series [5].
The key technical tool that they need for our analysis is the discrepancy measure that quantifies the divergence of the target and sample distributions defined by
\begin{center}\includegraphics[width=0.5\linewidth]{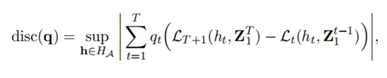}\end{center}

where \( q = (q_{1},...,q_{T})\)
is an arbitrary weight vector and where \(H_{A}\) is a set of sequences of hypotheses that the on-line algorithm A can pick. They were able to solve this constrained optimization problem by first solving a smaller convex optimization problem for \(q*\). Then, if \(q*>0\) they can guarantee that solving a kernel ridge regression problem for discrepancy is convex. Leading to their conclusion about discrepancy.  
The on-line learning scenario requires no distributional assumption.
In on-line learning, the sequence is revealed one observation at a time and it is often assumed to be generated in an adversarial fashion.
The goal of the learner in this scenario is to achieve a regret, that is the difference between the cumulative loss suffered and that of the best expert in hindsight, that grows sub-linearly with time.

\par \hspace{4ex} By using the OTB algorithm, we can convert our original models into dynamic ones, which will update the information every time we receive new data,and we can keep adjusting the expert weights to the most recent situation. 

\section{Data}

\par \hspace{4ex}Equity data was obtained from the Time Series Data API maintained by AlphaVantage [4]. Historical equity data is available at 4 time resolutions: intradaily, daily, weekly, and monthly. Our models were initially tested on daily closing prices of the S\&P 500 index fund and its constituents from 2000 to present. Various other stocks were used on a case-by-case basis.

\par \hspace{4ex} It should be noted that stock price is not continuous over time - the U.S. markets are open from 9 am to 4pm on business days, which excludes weekends and 10 holidays. 

\section{Model Performance}

\par \hspace{4ex}Empirical testing of the Linear Model, Generalized Linear Model, and Recurrent Neural Network, as compared to the martingale baseline, are detailed in the following sections.

\subsection{Linear Model}

\par \hspace{4ex}In testing, we found the linear model never beat the martingale, and performed worse when the "memory" of the model (number of time-lag points included) was increased. RMSE was used as an error metric. The tests below predict the price of SPX stock using previous day or days of SPX prices.

\par \hspace{4ex}The first test performed used the previous day's SPX value as the predictor and trained the model weights on the first 12 years of the data (2000 - 2011, approximately 70 percent of the data) and tested the model predictions on the last 5 years of the data (2012 - 2017, approximately 30 percent of the data). While a very naive test, this train/test split matches that of the RNN, for comparison purposes of comparison. 

\begin{figure}[h]
  \centering
  \includegraphics[width=0.5\linewidth]{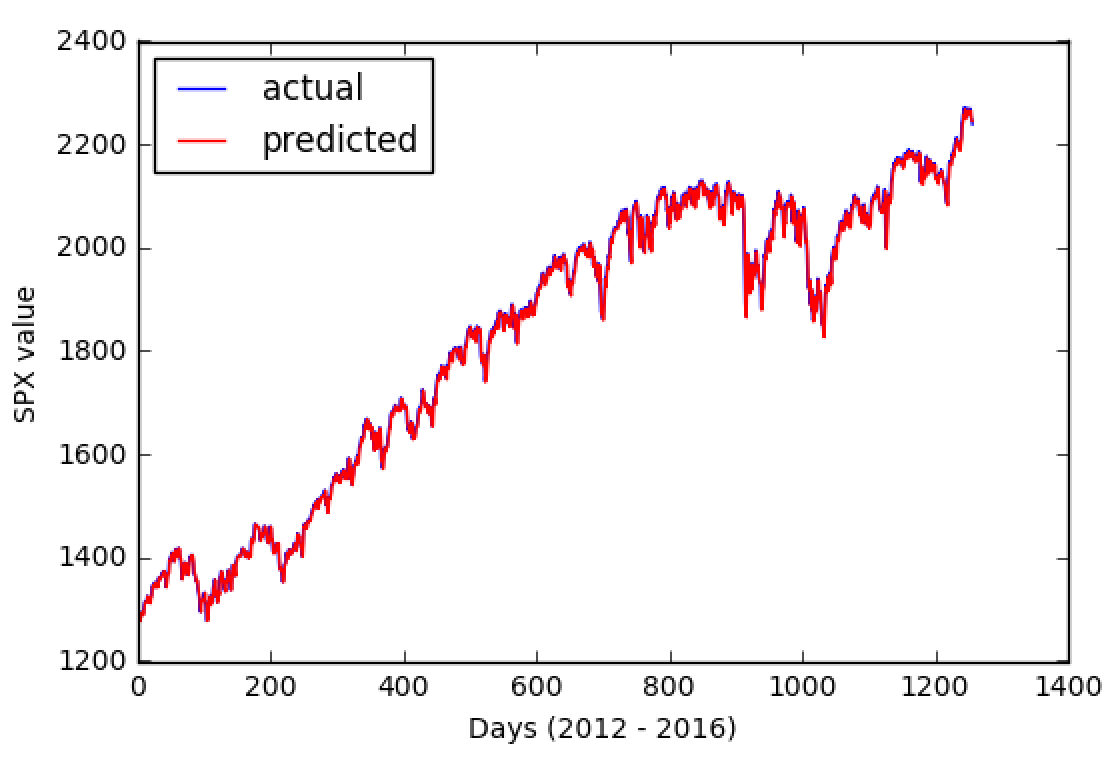}
  \caption{Performance of Baseline Linear Model}
\end{figure}

In this test, the model RMSE was 15.168, while the corresponding martingale RMSE was 14.867.

\par \hspace{4ex}The second test used the same train/test split of data, but used a set of SPX values at different time lags as the response variables. The time lags were chosen to incorporate values for 1-4 days prior, 1 week prior, approximately 1 month prior, and approximately 1 quarter prior, on the premise that modeling periodic trends could improve the model accuracy.

\begin{figure}[h]
  \centering
  \includegraphics[width=0.5\linewidth]{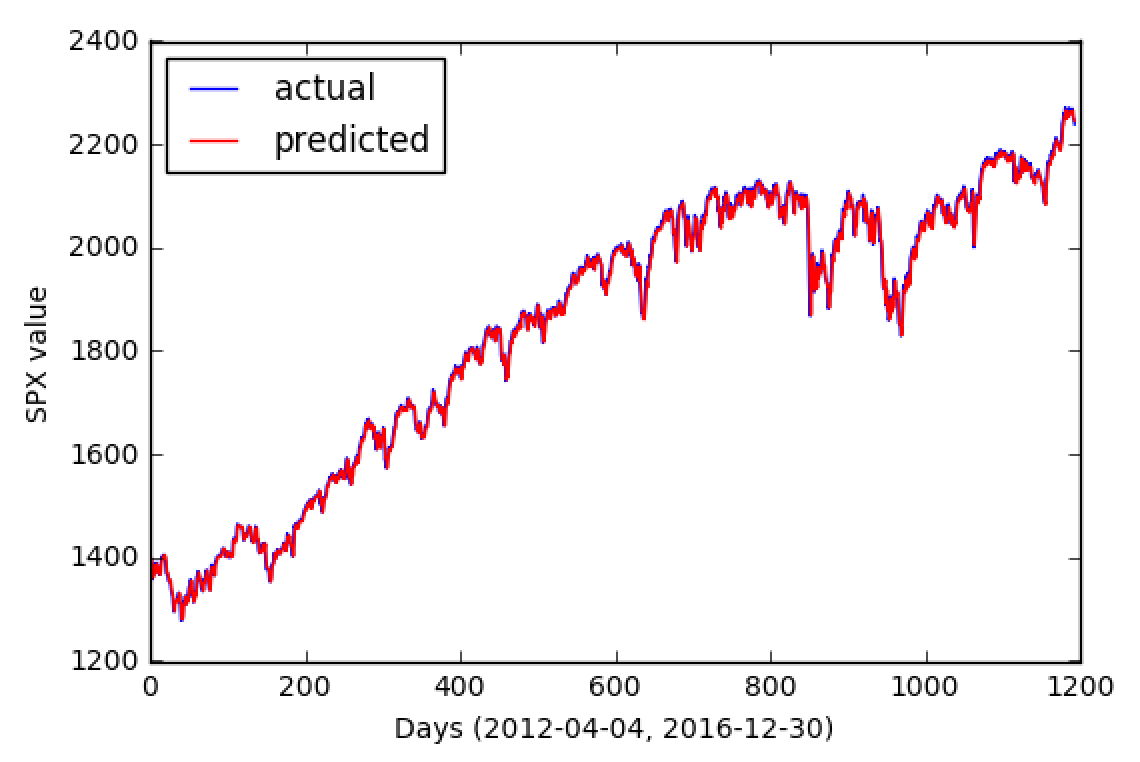}
  \caption{Performance of Linear Model with Lag}
\end{figure}

In this test, the model RMSE was 15.679 and the martingale RMSE was 15.141. (Note, the martingale RMSE is different from in Test 1 because the testing dates began in April, to allow the time lags to be used at all testing points).

From these results, we conclude that neither linear model outperformed the martingale model, and the inclusion of near-term periodic trends did not improve the model performance. 

\subsection{Generalized Linear Model}

\par \hspace{4ex}The GLM described in section 2.3 was first tested using the same 70/30 train/test data split, for a model predicting SXP closing price from the prior day's closing price. 

\begin{figure}[h]
  \centering
  \includegraphics[width=0.5\linewidth]{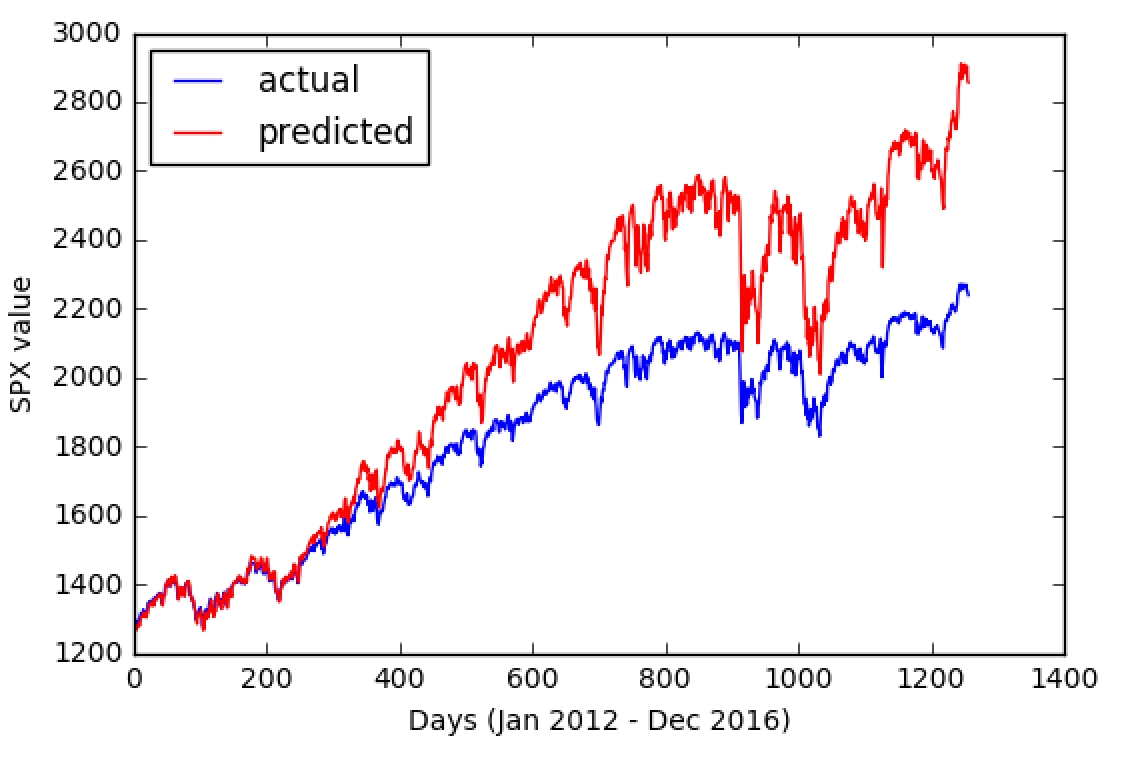}
  \caption{Prediction of SPX closing price from prior day's SPX price; GLM trained on data from 2000-2011.}
\end{figure}

The model predictions diverged greatly from the truth data (RMSE of 290.5), particularly as the test data exceeded the range of values seen in the training data. Such a change could not be captured by a GLM with stationarity assumptions.

\par \hspace{4ex}To address this, a second test was performed by breaking train/test data into one year segments, presuming that the GLM performance will improve with shorter training/testing periods. The year with the smallest RMSE is show below. 

\begin{figure}[h]
  \centering
  \includegraphics[width=0.5\linewidth]{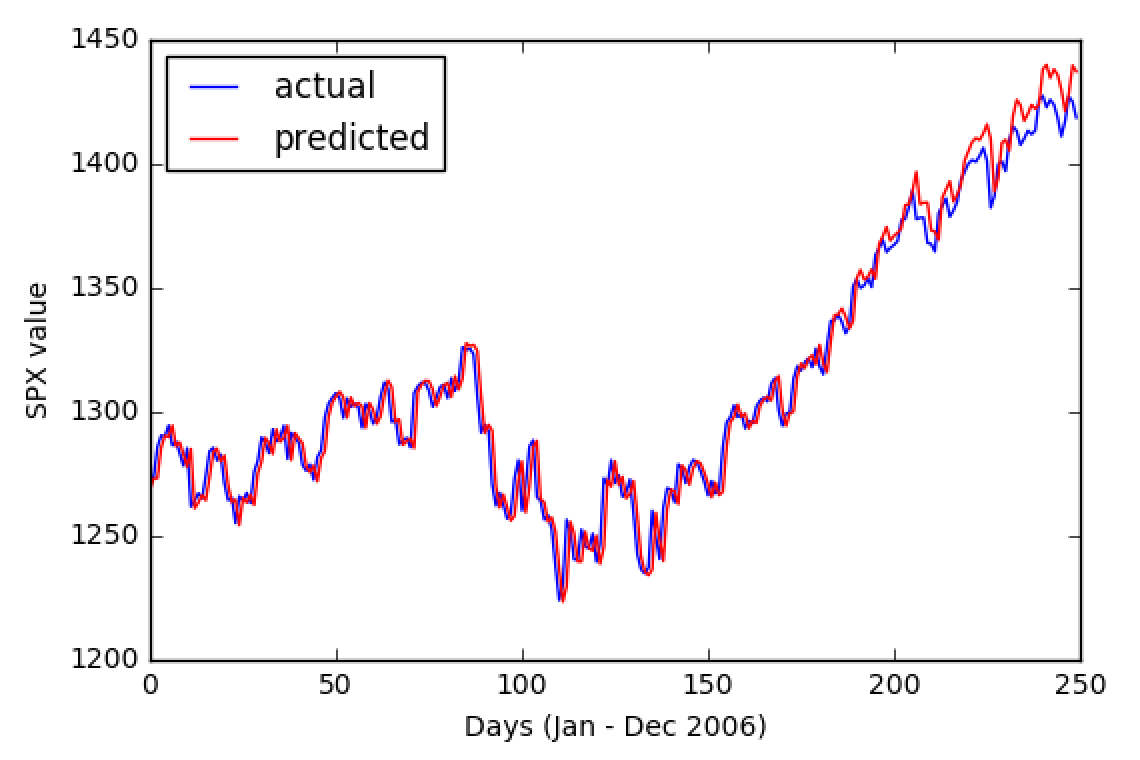}
  \caption{Prediction of SPX closing price from prior day's SPX price; GLM trained on data from 2005-2006.}
\end{figure}

Again, we find that the model diverges towards the end of the test data and that the model RMSE (8.91) is greater than the martingale RMSE (8.06). We find that one year is still too long of a period to go before updating weight - the GLM simply does not generalize over long time periods.

An online GLM implementation may have a better chance at beating or matching the martingale mode.

\subsection{Recurrent Neural Network}

\par \hspace{4ex}In testing, and refining our LSTM-RNN against the martingale model, we found that overall, the LSTM-RNN either learned the martingale, or did worse than the martingale. Many iterations were performed of different combinations of variables. For the untransformed S\&P 500 index-Daily data, 200 nodes performed the best on our testing set, with 21.43307 Mean Absolute Error (MAE), while the martingale performed twice as well with 10.53509 MAE.
\par \hspace{4ex}Next we tried to transform the data by division of the closing price by the opening price or percent change. This was meant to match the Geometric random walk model. Under this transform 10 LSTM nodes were tried and had moderately worse results than the martingale model.
We also tried to transform the data by subtraction of the closing price by the opening price, or absolute change. Matching the random walk model with Gaussian noise. Under this transformation 10 LSTM nodes were tried, and had as well moderately worse results than the martingale model.

\begin{table}[h]
  \caption{LSTM RNN results}
  \label{sample-table}
  \centering
  \begin{tabular}{lll}
    \toprule
    \multicolumn{3}{c}{}                   \\
    LSTM-RNN Model   &  mean L1-loss: LSTM    & mean L1-loss: martingale\\
    \midrule
    4 LSTM nodes     & 65.86085               & 10.53509      \\
    50 LSTM nodes    & 24.85500               & 10.53509   \\
    200 LSTM nodes   & 21.43307 			  & 10.53509   \\
    10 LSTM nodes, subtraction & 10.53782  & 10.53701         \\
    10 LSTM nodes, division  & 0.005847  & 0.005861       \\
    \bottomrule
  \end{tabular}
\end{table}

\begin{figure}[h]
  \centering
  \includegraphics[width=0.5\linewidth]{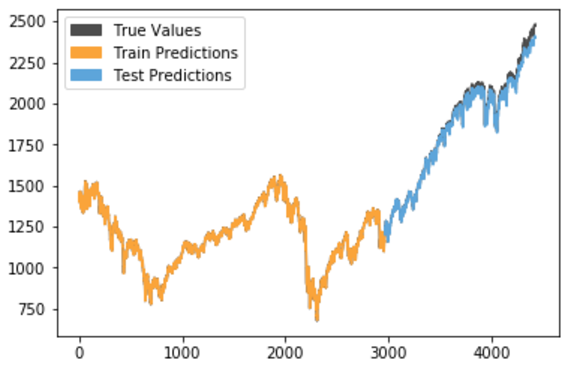}
  \caption{Performance of 200 LSTM nodes on SPX data}
\end{figure}

\par \hspace{4ex} In results on other stocks within the S\&P 500 index, we received better results whenever the testing data was within the range of the training data, such as for Suntrust Banks Inc. (STI). On STI, 4 nodes performed well with 0.59893 MAE versus the 0.57283 MAE of the Martingale model, the results of this test are in figure 6.

\begin{figure}[h]
  \centering
  \includegraphics[width=0.5\linewidth]{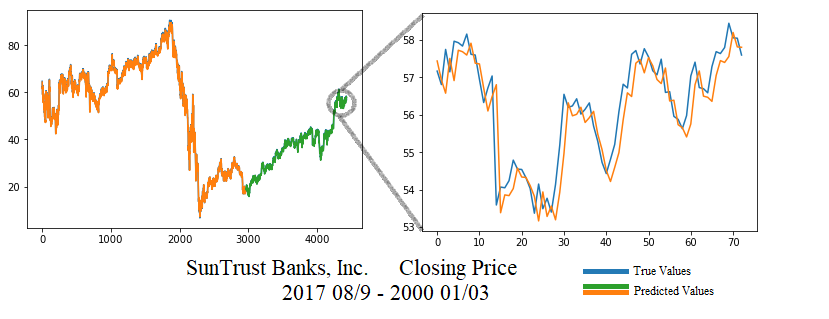}
  \caption{Performance of LSTM on Suntrust data}
\end{figure}

\par \hspace{4ex} As can be seen in figure 6. The LSTM lags behind the true value, and ends up just predicting whatever the previous value was. This was confirmed when the difference the two models was computed for this test to be 0.09230 MAE. Essentially, the neural network learned to compute the mean. This makes sense, as the mean is a generally good prediction, and an easy local minima for our model to fall into.

\par \hspace{4ex} In conclusion from our results on the LSTM-RNN. The RNN was not able to consistently beat the Martingale model, and even when transforms were placed on the data to force it to predict close to the Martingale, it always performed worse. This bolsters the idea that these stocks are correctly modeled by a martingale.
\section{Trading Strategies for Applying Stock Price Forecasting}

The following sections are techniques commonly used in algorithm trading and the financial industry. By using the models we mentioned before, if we can beat the martingale, we can make profits from the following trading strategies.

\subsection{Using call option and put options to make money}
\hspace{4ex} An option is a contract that gives its owner the right to buy (call option) or sell (put option) a financial asset (the underlying) at a fixed price (the strike price) at or before a fixed date (the expiry date). If you sell short (write) an option, you’re taking the other side of the trade. So you can enter a position in 4 different ways: buy a call, buy a put, sell short a call, sell short a put. And this with all possible combinations of strike prices and expiry dates. 

\par \hspace{4ex} The premium is the price that you pay or collect for buying or selling an option. It is far less than the price of the underlying stock. Major option markets are usually liquid, so you can anytime buy, write, or sell an option with any reasonable strike price and expiry date. If the current underlying price (the spot price) of a call option lies above the strike price, the option is in the money; otherwise it’s out of the money. 

\par \hspace{4ex} The opposite is true for put options. In-the-money is good for the buyer and bad for the seller. Options in the money can be exercised and are then exchanged for the underlying at the strike price. The difference of spot and strike is the buyer’s profit and the seller’s loss. {[8]} 
\begin{center}\includegraphics[width=0.6\linewidth]{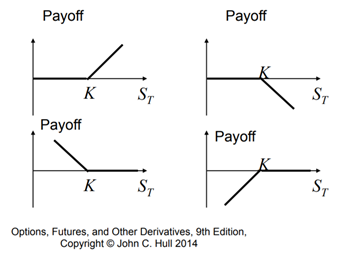}\end{center}

\par \hspace{4ex} Where K = strike price and St= stock price . Therefore, if we can minimize the loss function  in GLM and RNN, we’ll suffer lesser losses when exercising our put options and call options. Since the profit of execution after deducting cost of put and call options will still be greater or equal to zero. 

\subsection{Trading strategies}

\subsubsection{Arbitrage}

\par \hspace{4ex}Arbitrage is the difference of market prices between two different entities. Arbitrage is commonly practiced in global businesses. For example, companies are able to take advantage of cheaper supplies or labor from other countries. These companies are able to cut costs and increase profits. Arbitrage can also be utilized in trading S\&P  futures and the S\&P 500 stocks. It is typical for S\&P futures and S\&P 500 stocks to develop price differences. When this occurs, the stocks traded on the NASDAQ and NYSE markets either lag behind or get ahead of the S\&P futures, providing an opportunity for arbitrage. High-speed algorithmic trading can track these movements and profit from the price differences.

\subsubsection{Mean Reversion}

\par \hspace{4ex} Mean reversion is mathematical method that computes the average of a security's temporary high and low prices. Algorithmic trading computes this average and the potential profit from the movement of the security's price as it either goes away from or goes toward the mean price.

\subsubsection{Scalping}

\par \hspace{4ex}Scalpers profit from trading the bid-ask spread as fast as possible numerous times a day. Price movements must be less than the security's spread. These movements happen within minutes or less, thus the need for quick decisions, which can be optimized by algorithmic trading formulas.Other strategies optimized by algorithmic trading include transaction cost reduction and other strategies pertaining to dark pools.{[7]} In an online-to-batch algorithm design,this strategy can be easily done.

\subsection{A Win-win or Zero-sum game }
\par \hspace{4ex}No matter what trading strategy we use, a good algorithm can always minimize our losses without having extra costs to transaction, a win-win game therefore is possible even when we face structural change or sudden breakdown in market mechanism. 
On the other hand, with careful selection of put and call options portfolio, we can prevent from losing too much money from forecasting errors, which will eventually at least create a zero-sum situation in the long run. The most important is that: this trading algorithm and strategy is not hard to implement and utilize by laymen or regular end users.

\section{Conclusions}
\hspace{4ex}The aim of this research is to develop a predictive model to forecast financial time series data. In this study, we have examined 5 and developed 4 predictive models. The mean and linear regression analysis imply that the predictive values and the real values are deviating from the mean. Then we take the GLM and RNN model compared with Mean and ordinary linear model. Empirical examinations of predicting precision for the price time series (by the comparisons of predicting measures as MAE and RMSE) show that the proposed models (GLM, LSTM-RNN) fail to improve on the precision of forecasting 1 dimensional time series. Further improvement with state of the art techniques could be used with trading strategies mentioned in section 5 to capitalize on stock price forecasts.

\section*{References}
\small

[1]\ Shreve, Steven E. Stochastic Calculus for Finance I The Binomial Asset Pricing Model. Springer, 2005.Alexander, J.A. 

[2]\ Pinsky, Mark A., and Samuel Karlin. An Introduction to Stochastic Modeling. Academic Press, 2011.

[3]\ Vitaly Kuznetsov and Mehryar Mohri. {\it Time series prediction and online learning.} In Proceedings of The 29th Annual Conference on Learning Theory (COLT 2016). New York, USA, June 2016. 

[4]\ https://www.alphavantage.co/documentation/ 

[5]\ {\it Introduction to Generalized Linear Models.} Penn State, Eberly College of Science.  https://onlinecourses.science.psu.edu/stat504/node/216

[6]\ http://www.statsmodels.org

[7]\ Hull, J. (2006). Options, futures, and other derivatives. Upper Saddle River, N.J: Pearson/Prentice Hall.

[8]\ http://www.investopedia.com/terms/a/algorithmictrading.asp

[9]\ Algorithmic Options Trading, Part 1 http://www.financial-hacker.com/algorithmic-options-trading/

\end{document}